\documentclass[10pt,a4paper]{article}
\usepackage{hybridreport}
\usepackage[square,authoryear]{natbib}
\usepackage{multirow}
\usepackage{array}
\newenvironment{acknowledgements}{\section*{Acknowledgment}}{}

\renewcommand{\thesection}{\Roman{section}}
\renewcommand{\thesubsection}{\thesection-\Alph{subsection}}

\reporttype{Technical Report}
\logo{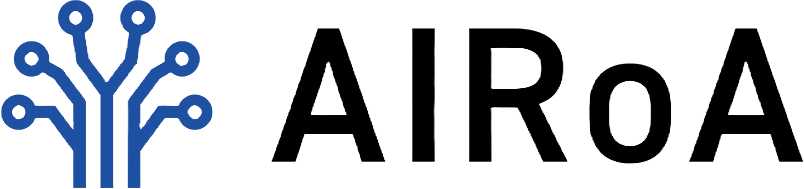}
\title{YUBI-Real2Sim: 3DGS Scene Verification for Reliable Policy Evaluation}
\shorttitle{YUBI-Real2Sim: 3DGS Scene Verification for Reliable Policy Evaluation}
\author{%
\mbox{Hiroki Nishizawa\textsuperscript{1,2}}\quad
\mbox{Takehiko Ohkawa\textsuperscript{1}}\quad
\mbox{Nicola Scianca\textsuperscript{1}}\\
\mbox{Tatsuya Matsushima\textsuperscript{1}}\quad
\mbox{Shigeo Morishima\textsuperscript{2}}\quad
\mbox{Kei Ota\textsuperscript{1}}}
\institution{\textsuperscript{1}AI Robot Association (AIRoA)\quad
\textsuperscript{2}Waseda University}
\date{October 2026}
\hypersetup{pdftitle={YUBI-Real2Sim: 3DGS Scene Verification for Reliable Policy Evaluation},
pdfauthor={Hiroki Nishizawa, Takehiko Ohkawa, Nicola Scianca, Tatsuya Matsushima, Shigeo Morishima, Kei Ota}}
\abstracttext{

Evaluating VLA policies on real robots is costly, time-consuming, and difficult to reproduce, making real-to-sim simulation essential for scalable evaluation.
While 3D Gaussian Splatting (3DGS) enables rapid real-to-sim scene generation, the reconstruction quality required for reliable policy evaluation remains unquantified.
In particular, policies relying on wrist-mounted fisheye cameras (e.g., YUBI handheld gripper data) are heavily exposed to visual artifacts (holes, floaters) in wide-angle views, directly degrading task performance.
To address this, we present \emph{YUBI-Real2Sim}, a unified framework for 3DGS-based scene reconstruction and real-to-sim policy evaluation under dual fisheye wrist views.
We develop an integrated pipeline combining 200-degree fisheye camera capture, 3DGS training, 3D foundation models for object modeling, and MuJoCo-based physical simulation, reproducing the observed real-world policy ordering in eight of nine scene--task comparisons.
Using this framework, we construct a benchmark across two bimanual robot platforms,
three scenes, three tasks, and dual $\pi_{0.5}$-based policies.
This systematically measures real-to-sim rank agreement and evaluates policy sensitivity against counterfactual 3DGS artifacts (ceiling holes, floaters, warping) and capture disturbances.
We further analyze whether novel-view quality scores can indicate policy-ranking reliability under a scene-held-out evaluation protocol. The results show task-dependent performance and limitations of quality-based reliability prediction, highlighting the need to assess reconstruction quality when using Real2Sim for policy evaluation. These results do not yet establish a generally reliable scene-acceptance rule.
Project Page: \url{https://yubi-real2sim.airoa.io}.

}
\begin{document}
\setbox0\hbox{\includegraphics[height=14pt]{figs/logo.png}}
\maketitle
\begin{figure}[H]

\centering
\includegraphics[width=\textwidth]{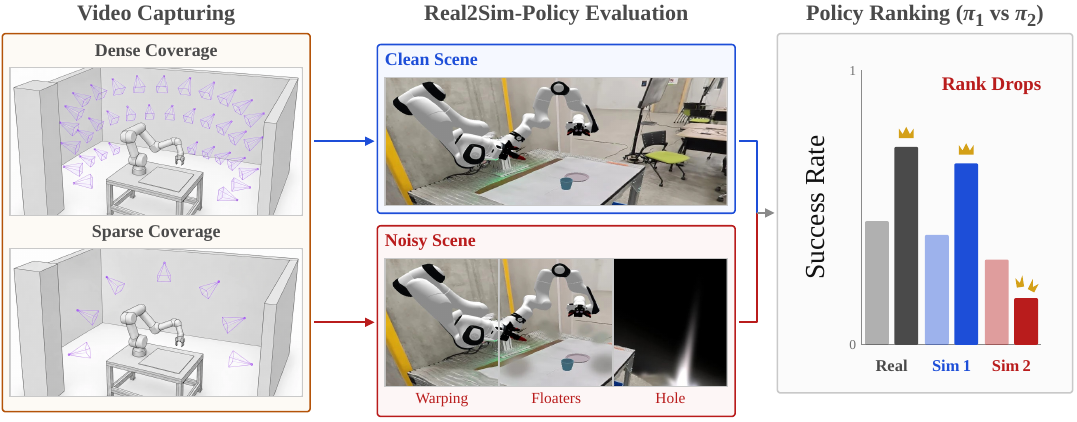}
\caption{\textbf{YUBI-Real2Sim analyzes and verifies 3DGS scenes for reliable policy-ranking evaluation.} The quality of 3DGS can easily degrade depending on the camera views. We analyze how variations in 3DGS quality affect policy rankings
and examine whether scene-quality scores can indicate
policy-ranking reliability.}
\label{fig:teaser}

\end{figure}

\section{Introduction}
\label{sec:introduction}

Evaluating VLA policies requires repeated robot
execution and physical resets, limiting the policies and environments that
can be compared~\cite{jin2026umibench}. A practical use of simulation is to
choose which of two candidate policies to deploy. For that decision, the
relevant outcome is whether simulation preserves their real-world policy ranking.

Real-to-sim systems reconstruct environments from real captures and use
them for policy evaluation~\cite{jain2025polaris,ranawaka2026simfoundry}.
3D Gaussian Splatting (3DGS)~\cite{kerbl2023gs} makes this approach visually
expressive, while SIMPLER~\cite{li2024simpler} motivates judging a simulator
by the policy comparisons it supports. However, a reconstructed background
may contain holes, floaters, and warped surfaces in regions poorly observed
during capture. PolaRiS uses 2D Gaussian Splatting and co-trains policies on real and simulated data to improve policy-ranking correlation~\cite{jain2025polaris}.

This issue is particularly relevant to YUBI~\cite{ohkawa2026yubi}, a
UMI-style~\cite{chi2024umi} interface for bimanual demonstration collection.
YUBI policies observe the environment through two independently moving
fisheye wrist cameras. Their wide-angle observations include regions beyond
the immediate manipulation target, and their viewpoints change throughout
execution. Evaluating these policies therefore requires both a suitable
visual simulator and evidence that its scene reconstruction supports reliable
policy selection.

We present \emph{YUBI-Real2Sim}, a benchmark and reconstruction pipeline for
this setting. It covers Duo and Flat on a dual Franka FR3 setup and the TX-G2
scene on a TX-G2 (AgiBot-G2-compatible) platform, written \emph{TX-G2}
throughout. Each task supplies two
$\pi_{0.5}$-based candidate policies~\cite{black2025pi05}; one
scene--task pair contributes one binary policy comparison.

First we show the clean reconstruction from our pipeline achieves \textbf{88.9\%}
    agreement with the real policy rankings in three scenes with three tasks. We then use the clean reconstruction as a reference and edit only the
background Gaussians. Ceiling holes, floaters, and warping
test whether visual defects cause previously correct policy comparisons to
become reversed or tied. Reduced-coverage reconstructions serve a supporting
role: they test whether analogous visible defects arise when the input
images provide less support. Finally, we evaluate NVS-SQA~\cite{qu2026nvssqa}
as a candidate indicator of ranking
reliability, with separate validation required for a scene-rejection rule.

Our contributions are:
\begin{itemize}
    \item \textbf{A Real2Sim pipeline for YUBI policy evaluation.}
    We integrate scene reconstruction and dual-fisheye policy evaluation on the dual Franka FR3 and TX-G2 platforms. We will release scenes, environments, evaluation code.

    \item \textbf{A benchmark for YUBI policy evaluation in Real2Sim.}
    We evaluate two $\pi_{0.5}$-based candidate policies per task across three scenes and three tasks: Cup, Tape, and Pen. Clean simulation reproduces the observed real policy ranking in 8 of 9 scene--task comparisons (88.9\%).

    \item \textbf{An analysis of reconstruction defects and policy rankings.}
    We analyze how counterfactual 3DGS artifacts affect policy rankings under fixed physics, objects, and rollout conditions. We also examine reduced-coverage reconstructions to study defects arising from limited image support.

    \item \textbf{An analysis of image-quality metrics for policy evaluation.}
We analyze the relationship between reconstruction quality and real-to-sim policy-ranking agreement using standard image-quality metrics (PSNR, SSIM~\cite{wang2004ssim}, and LPIPS~\cite{zhang2018lpips}), alongside NVS-SQA.
\end{itemize}

\section{Related Work}
\label{sec:related}

\textbf{Real-to-sim policy evaluation.}
SIMPLER~\cite{li2024simpler} evaluates whether visually matched simulation
preserves policy performance comparisons, using success-rate correlation
and mean maximum rank violation (MMRV). PolaRiS~\cite{jain2025polaris} and
SimFoundry~\cite{ranawaka2026simfoundry} extend evaluation to reconstructed
scenes, including mesh objects and Gaussian-splat backgrounds. We focus on one
specific decision: selecting the better of two policies for the same task
and scene, measured as the fraction of correctly reproduced policy rankings.
This pairwise accuracy is distinct from success-rate correlation and MMRV.
Related systems reconstruct scenes for policy learning or failure
discovery~\cite{torne2024rialto,han2025re3sim,jiang2025gsworld,zhang2025real2simeval,zhang2026robosnap,sagar2024robomd},
while SureSim~\cite{suresim2025} corrects an imperfect simulator using paired
real trials. Our experimental variable is the visual reconstruction: we
measure whether changing it alters which policy simulation selects.

\textbf{Evaluating novel-view synthesis.}
Existing reconstruction-based real-to-sim systems evaluate policy performance
in reconstructed scenes~\cite{jain2025polaris,ranawaka2026simfoundry}, but a
remaining gap is a validated reconstruction-quality metric for identifying
scenes that support reliable policy comparisons. Novel-view synthesis (NVS)
provides candidate quality measures, although their relevance depends on the
viewpoints being evaluated. Every-$N$th-frame holdouts mainly measure
interpolation; spatial-sector holdouts reveal a 3--12\,dB
interpolation--extrapolation gap that can reverse reconstruction-method
rankings~\cite{jia2026mind}. This matters for YUBI, whose independently
moving fisheye wrist cameras may observe viewpoints poorly supported by the
capture trajectory. When aligned ground-truth views are unavailable,
NVS-SQA~\cite{qu2026nvssqa} scores the perceptual quality of rendered sequences
without reference images, while CrossScore~\cite{wang2024crossscore} predicts
SSIM using unaligned neighboring captures. Fisheye-specific Gaussian
Splatting~\cite{liao2024fisheyegs,deng2025selfcaligs,huang2025scomnigs} addresses
wide-field rendering, but rendering quality alone does not establish ranking
reliability. We therefore use NVS metrics as candidate diagnostics and test
whether their scores identify reconstructed scene variants that fail to
preserve real-world pairwise policy rankings.
\begin{figure*}[t]
\centering
\includegraphics[width=\textwidth]{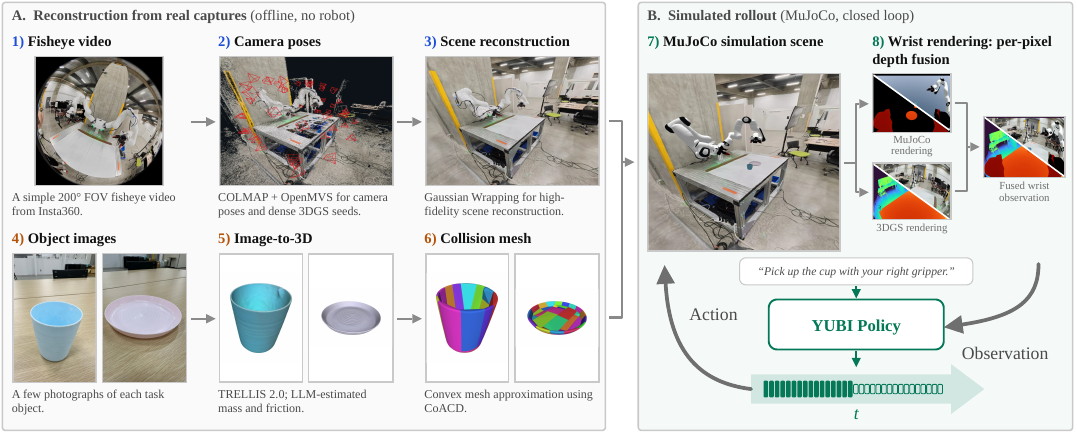}
\caption{\textbf{YUBI-Real2Sim pipeline.} We capture the background with a single $200^\circ$ FOV camera, estimate camera poses with COLMAP, and densify the point cloud with OpenMVS to initialize Gaussian Wrapping for high-fidelity reconstruction. We generate object geometry and textures with TRELLIS 2.0 and approximate collision meshes with CoACD. The reconstructed scene and objects are integrated into MuJoCo simulation. For wrist-view rendering, we independently render the background Gaussians and the MuJoCo visual meshes, then composite their pixels using depth maps.}
\label{fig:pipeline}
\end{figure*}
\section{YUBI-Real2Sim System}
\label{sec:system}

YUBI-Real2Sim integrates environment capture, static-scene reconstruction,
object modeling, and closed-loop policy evaluation under dual fisheye wrist
observations (Fig.~\ref{fig:pipeline}). We project selected fisheye frames
into pinhole views, estimate camera poses with COLMAP, generate a dense seed
point cloud with OpenMVS, and optimize a 3DGS background with Gaussian
Wrapping. Task objects are modeled separately as visual and collision meshes
and assembled with the robot in MuJoCo, with the reconstructed background
registered to the simulation coordinate system. During simulation, background
and mesh renders are composited by depth to form the two calibrated YUBI
wrist-camera images. A policy receives these images and proprioception and
returns action chunks that drive the simulated robot, closing the loop with
the observation modalities and action interface used on hardware.
Fig.~\ref{fig:pipeline} summarizes the system.

\textbf{Evaluation setting.}
We consider three scenes on two bimanual platforms. \emph{Duo} and
\emph{Flat} are scenes on a dual Franka FR3 setup and differ in their table
and background. The third scene uses a TX-G2 (AgiBot-G2-compatible) platform
and is denoted \emph{TX-G2}. Both platforms carry a YUBI gripper with a fisheye
camera on each wrist. The task inventory comprises three tasks: Cup (cup
pick-and-place), Tape (placing tape into a box), and Pen (placing a pen into a
box). The current ranking results use all three tasks; evaluation details
appear in Sec.~\ref{sec:benchmark}.

\textbf{Capture.}
Reconstructing the environment for two independently moving fisheye wrist
cameras requires visual coverage beyond the immediate manipulation target.
Narrow-angle capture can leave surrounding regions poorly observed even
though they later appear in policy inputs. We therefore use an Insta360
camera to obtain broad angular coverage with a compact source-frame set,
keeping reconstruction optimization practical. Reconstruction uses only its
left fisheye lens (\texttt{Insta360 v6} in the camera metadata).
Videos contain two $3840\times3840$ tracks at 50\,fps (HEVC). Capture durations
are 213.06\,s (Duo), 254.88\,s (Flat), and 243.50\,s (TX-G2).
We partition each video temporally and select one source frame per interval
that maximizes the minimum left/right Laplacian variance, without an explicit
manual blur-rejection threshold. The same source timestamps also support a
controlled comparison between the full fisheye-derived projections and a
central pinhole view (Sec.~\ref{sec:disturbance}).

\textbf{Reconstruction.}
Using OpenCV fisheye calibration, each source frame is reprojected into nine
$2048\times2048$ pinhole views with $90^\circ\times90^\circ$ FOV and focal
length 1024\,px: one forward view and eight views tilted $55^\circ$ from
forward, spaced at $45^\circ$ azimuth intervals.
We estimate poses with COLMAP~\cite{schoenberger2016sfm,schoenberger2016mvs} through pycolmap 4.1.1, generate a dense seed
point cloud with OpenMVS \cite{openmvs}, and train Gaussian
Wrapping~\cite{gomez2026gaussianwrapping} at full resolution for 30,000--60,000 iterations using an NVIDIA H200 GPU.
The initial dense seed is capped at 40 million points with seed 0.
Exposure compensation, normal fields, fast multiview regularization, and
masked rendering loss are enabled.
The single-lens reconstruction has arbitrary scale. A stored transformation
registers it to MuJoCo world coordinates. All embodiments
are manually aligned to robot visual meshes, and the tabletop collision
surface is modeled as the plane $z=0$.
Collision OBJ meshes are supplied separately from the 3DGS training path.
Task objects are reconstructed as separate meshes with TRELLIS~2.0
\cite{xiang2025trellis2} and given CoACD~\cite{wei2022coacd} collision, placed at the
real initial layout, and given LLM-estimated mass and friction.

\textbf{Simulation and rendering.}
Physics runs in MuJoCo~\cite{todorov2012mujoco} in an environment derived from
robosuite~\cite{zhu2020robosuite}, with the robot and objects rendered as
meshes and the static background rendered from the 3DGS, composited per pixel
by depth.
Each YUBI wrist camera is rendered as a cubemap from the 3DGS and remapped to
the fisheye image with the intrinsics calibrated on the real YUBI camera
(\texttt{ELP-USB16MP01-L18}), at the resolution the policy consumes on the real robot.
The camera positions and orientations relative to the wrists are manually
fine-tuned in simulation with reference to the physical camera mounts.
For each robot configuration, the wrist-relative camera poses are held fixed
across the three benchmark tasks, the two candidate policies for each task,
and all scene perturbations. This keeps the rendered viewpoint and camera
calibration constant, so comparisons across policies and scene variants are
not confounded by camera-pose changes.
To avoid gradual contact slip inherent in MuJoCo's soft-contact
model~\cite{todorov2012mujoco,mujoco_slip_documentation}, grasps are realized by a contact rule:
an object is attached to a gripper when both of its finger inner surfaces
touch the object.
This attachment is a modeling approximation that bypasses force-dependent
grasp stability and suppresses slip-related grasp failures.

\textbf{Policy interface.}
$\pi_{0.5}$~\cite{black2025pi05} policies receive the two fisheye wrist images
and proprioception and predict end-effector delta actions referenced to the
left hand, together with gripper opening. Actions are predicted in chunks of
32, of which 16 (Cup) or 8 (Tape and Pen) are executed per chunk on both the
real robot and in simulation.
The FR3 hardware uses impedance control. The TX-G2 setup uses PID control with
Ruckig~\cite{berscheid2021ruckig} on both the real robot and in simulation.

\section{Benchmark Protocol}
\label{sec:benchmark}

\textbf{Scenes and platforms.}
The benchmark targets three scenes: \emph{Duo} and \emph{Flat} on a dual
Franka FR3 setup, and \emph{TX-G2} on a TX-G2 (AgiBot-G2-compatible) platform.
Duo and Flat differ in table and background.
Both platforms carry YUBI grippers on both wrists. The same policy runs on
both through platform-specific action mappings and low-level controllers.

\textbf{Tasks.}
The benchmark contains three tasks: Cup (cup pick-and-place), Tape (placing
tape into a box), and Pen (placing a pen into a box). These are the only tasks
included in the ranking analysis. A trial is counted as successful as soon as the task-specific
success criterion is met within the time limit; otherwise,
it is counted as a failure.
The time limit is 3\,min (1{,}800 frames) for Tape and Pen
in Duo and for Pen in TX-G2, and 1\,min (600 frames) for all other scene--task pairs,
with actions executed at 10\,Hz.
Within each scene--task pair, the same time limit applies
to both policies in real evaluation and across all simulated
reconstruction conditions. For real
trials, an operator assigns the label using a fixed task-level rubric; for
simulation, a deterministic object-state predicate implements the corresponding
endpoint. The rubric and simulation predicate are held fixed across policies
within each task.

\textbf{Policies and counting unit.}
Each task has two task-specific $\pi_{0.5}$-based candidates, denoted
$\pi_1$ and $\pi_2$
with labels fixed before comparison. They are not two shared checkpoints
across tasks. Three tasks on three scenes provide 18 policy cells and nine
potential pairwise comparisons, all of which are evaluated in the current
result.

\textbf{Trials.}
Each policy cell uses 20 real and 80 simulated rollouts.
Within each scene--task comparison, both policies and the clean and
perturbed variants share the initial layout and initial-condition/seed schedule. Objects are reset to the same nominal layout before each trial.
Object assets and the simulated layout are matched to the corresponding real
setup. Simulated rollouts use different seeds for VLA inference, with the
same seed list reused across clean and perturbed scenes
as specified in Sec.~\ref{sec:conditions}.
\begin{figure*}[!t]
\centering
\includegraphics[width=\textwidth]{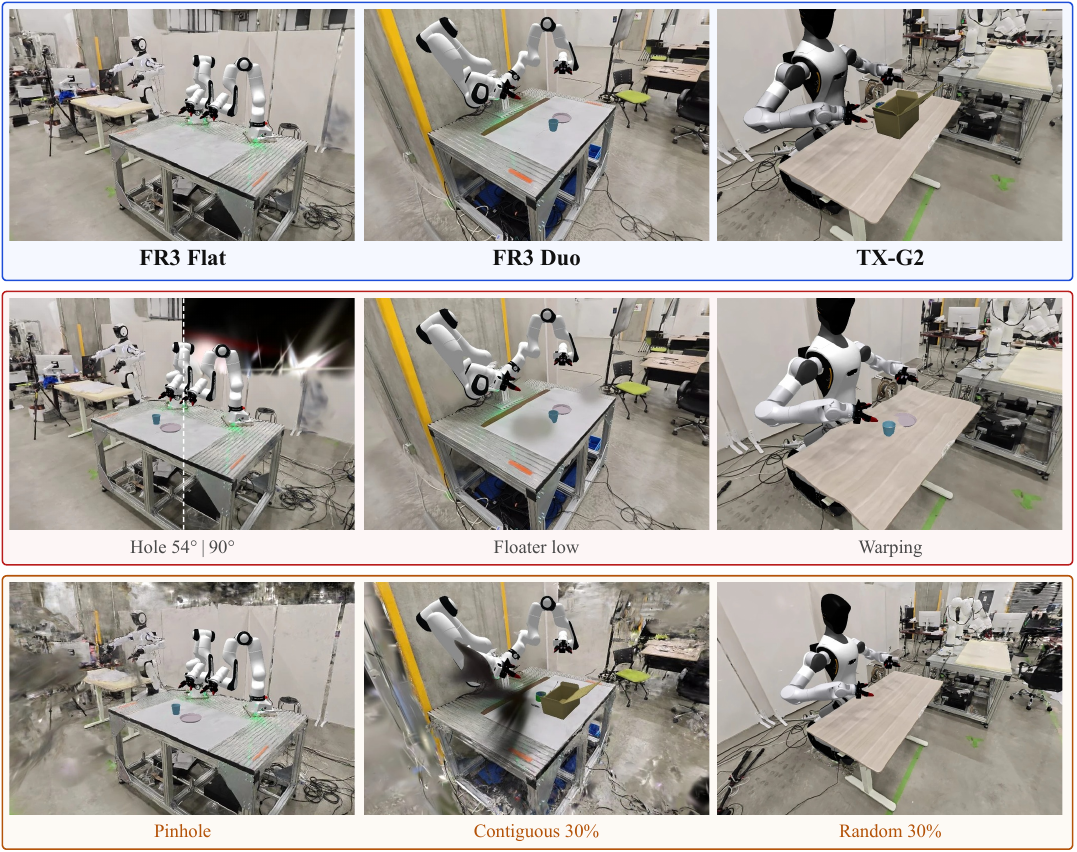}
\caption{\textbf{Simulated scenes, and the added disturbances for 3DGS or viewpoints.} One column per scene, one band per condition.
\textcolor{blue}{Blue}: the clean 3DGS reconstructions the rollouts run in.
\textcolor{red}{Red}: the same reconstructions carrying a counterfactual
artifact --- a ceiling hole in FR3 Flat, the $54^\circ$ and $90^\circ$
half-angles either side of the dashed rule, low floaters in FR3 Duo, and
warping in TX-G2. \textcolor[HTML]{B45309}{Amber}, the colour
Fig.~\ref{fig:teaser} gives capture: one reduced-coverage reconstruction per
condition, each in the scene of its column. The panels show representative examples of the reconstruction conditions. We evaluated two task-specific VLA checkpoints for each of three tasks (Cup, Tape, and Pen) in all three scenes (FR3 Flat, FR3 Duo, and TX-G2).}
\label{fig:analysis}
\end{figure*}
\subsection{Evaluation conditions}
\label{sec:conditions}
The evaluation has a clean reconstruction condition and two types of
counterfactual condition.  The artifact conditions edit the reconstructed
background while holding the robot, physics, collision geometry, object
assets, initial conditions, policy checkpoints, success rules, and rollout
seeds fixed.  The reduced-coverage conditions retrain the 3DGS scene with
fewer source images and are used for visual defect, quality, and policy-ranking
analyses.

\textbf{Clean and artifact conditions.}
The clean reconstruction uses all nine projected views per selected left-lens
source frame.  The artifact design contains six variants per scene: two
holes, three floater configurations, and one warping realization.  Each edit
changes only background Gaussians, leaving collision geometry and object
assets fixed.  Initial conditions, policy checkpoints, success rules, and
rollout seed lists are paired with the clean evaluation.

\emph{Hole}: an upward cone removes Gaussians at half-angles of $54^\circ$
and $90^\circ$ from the vertical.  The table protection region uses its
collision footprint expanded by 40\,mm, a $\pm60$\,mm slab about its surface,
and below-table protection; the ceiling directly above the table remains
eligible for removal.

\emph{Floaters}: eight clusters of four Gaussians with neighboring scene
colors are placed over the table in each of three height bands: 5--15,
15--25, and 25--35\,cm.  Placement maintains 10\,mm clearance from the
table and objects.  Within-cluster positional standard deviation is 18\,mm,
the Gaussian size center is 31.6\,mm, and the opacity center is 0.99;
view-dependent SH is disabled.  These are three artifact configurations;
isolating height alone additionally requires matching their horizontal
positions, sizes, and colors.

\emph{Warping}: one fixed seed places 30 smooth domes that displace
table-region Gaussians vertically, with sampled heights of 10--30\,mm and
radii of 30--60\,mm.  The region uses the footprint expanded by 40\,mm and a
$\pm60$\,mm surface band.  Overlapping displacements add without a cap, so
the total displacement may exceed 30\,mm.  One seed characterizes one fixed
realization rather than a distribution over random table defects.

\textbf{Reduced-coverage conditions.}
The supporting reconstruction study compares Clean, Central pinhole,
Contiguous 30\% removal, and Random 30\% removal reconstructions at the same held-out
and robot-view protocols.  Central pinhole retains only the forward left-lens
projection ($L_f$) at every source timestamp. Both removal conditions also
use only $L_f$: Contiguous 30\% removal drops one consecutive interval containing
30\% of the time-ordered selected source frames, and Random 30\% removal drops
the same number at random.  The
interval location and random seed are fixed before examining reconstructed
artifacts.

For a scene with $N$ source frames, the removal conditions use
\begin{equation}
 n_{\mathrm{remove}}=\operatorname{round}(0.30N),\qquad
 N_{\mathrm{keep}}=N-n_{\mathrm{remove}}.
 \label{eq:view-removal}
\end{equation}
Removal is applied to the central-pinhole source-frame set. This leaves
134/192 frames for Duo, 105/150 for Flat, and 153/219 for TX-G2.  These counts
precede splitting; quality-evaluation views are held fixed and excluded from
every training set.  The central-view condition changes projections per
source frame from nine to one; each projection remains $2048\times2048$ with
a $90^\circ\times90^\circ$ FOV.

COLMAP intrinsics, registered poses, and the original dense seed are reused.
Each condition uses fresh optimization with seed 0 and 30,000 iterations.
The same held-out views are used across all
reconstruction conditions and excluded from training.
Clean, edited, and reduced-coverage reconstructions are compared
at the same camera poses, including wrist views, and visible defect type,
location, and extent are documented.  Visual similarity alone does not
establish a policy-ranking effect for a retrained variant.

\subsection{Metrics and reporting}
\label{sec:metrics}

For scene $s$, task $t$, policy $p\in\{\pi_1,\pi_2\}$, and domain
$d\in\{\mathrm{real},\mathrm{sim}\}$, the success-rate estimate is
\begin{equation}
 \hat q_{s,t,p}^{d}=\frac{S_{s,t,p}^{d}}{N_{s,t,p}^{d}}.
 \label{eq:success-rate}
\end{equation}

\textbf{Pairwise ranking accuracy.}
For scene--task pair $i$, let $r_{i,\pi_1},r_{i,\pi_2}$ and
$s_{i,\pi_1},s_{i,\pi_2}$ be
observed real and simulated success rates. Define
\begin{equation}
 d_i^{\mathrm{real}}=r_{i,\pi_1}-r_{i,\pi_2},\qquad
 d_i^{\mathrm{sim}}=s_{i,\pi_1}-s_{i,\pi_2}.
\end{equation}
Let $M$ contain pairs with both required evaluations and a strict
observed real ranking ($d_i^{\mathrm{real}}\ne0$). Our primary metric is
\begin{equation}
 \mathrm{Acc}_{\mathrm{rank}} =
 \frac{1}{|M|}\sum_{i\in M}
 \mathbf{1}\!\left[d_i^{\mathrm{real}}d_i^{\mathrm{sim}}>0\right].
 \label{eq:rank-accuracy}
\end{equation}
The numerator is the number $k$ of eligible comparisons for which simulation
selects the same policy as the real evaluation, so
$\mathrm{Acc}_{\mathrm{rank}}=k/|M|$.  Thus this metric measures agreement in
the two-policy choice, not agreement in absolute success rates.
A reversed order and a simulation tie both score zero; we report them
separately. Real ties are unrankable and are reported separately. The
comparison set is fixed before testing scene perturbations, so a variant
with two zero-success policies remains in the denominator as a simulation
tie once that comparison is completed. We report correct/total pairs per scene
and overall, weighting each
eligible scene--task pair equally. Success counts support the ranking;
matching their absolute values is outside this benchmark's claim.

The success-rate bars show marginal 95\% Wilson score intervals using
20 real or 80 simulated trials per policy. These intervals quantify each
success-rate estimate; they are not confidence intervals for ranking accuracy
or tests of the difference between policies.
Ranking accuracy is reported as a point estimate over the fixed set of
eligible scene--task comparisons; rollout counts are not treated as
independent ranking comparisons.

We also report the absolute-rate error,
\begin{equation}
 \mathrm{MAE}=
 \frac{1}{2|M|}\sum_{i\in M}\sum_{p\in\{\pi_1,\pi_2\}}
 \left|\hat q_{i,p}^{\mathrm{real}}-\hat q_{i,p}^{\mathrm{sim}}\right|.
 \label{eq:rate-mae}
\end{equation}
This secondary metric does not determine whether the policy ranking is
correct. Only the Clean MAE is reported (Sec.~\ref{sec:clean-results}).

\textbf{Counterfactual comparison.}
For artifact variant $v$, we report its ranking-accuracy change from clean as
\begin{equation}
 \Delta\mathrm{Acc}_{\mathrm{rank}}(v)=
 \mathrm{Acc}_{\mathrm{rank}}(v)-\mathrm{Acc}_{\mathrm{rank}}(\mathrm{clean}).
 \label{eq:cf-rank-delta}
\end{equation}
We additionally count clean-correct comparisons that become reversed or tied,
and clean-incorrect comparisons that become correct. A decrease in both
policies' success rates is a rate change only; it is a ranking failure when
the policy ranking changes or becomes tied. Table~\ref{tab:cf-significance-ranking-transitions} reports these counts over the nine comparisons of each condition.

\textbf{Quality metrics and view protocols.}
For metric $m$, scene $s$, and view protocol $u$, we report the paired score
change
\begin{equation}
 \Delta Q_{m,s,u}=Q_{m,s,u}^{\mathrm{counterfactual}}
 -Q_{m,s,u}^{\mathrm{clean}}.
 \label{eq:quality-delta}
\end{equation}
NVS-SQA~\cite{qu2026nvssqa}
is evaluated on held-out Test views and OBS views. Test views have aligned
captured images, so PSNR, SSIM~\cite{wang2004ssim}, and LPIPS~\cite{zhang2018lpips} are also reported there; OBS has no
paired ground-truth image. OBS--Probe uses a fixed pre-rollout trajectory,
whereas OBS--Policy uses frames observed during policy rollouts. Scores are averaged over frames within each scene or
trajectory.

\textbf{Scene verification without target-scene real evaluation.}
For task $t$ in scene variant $(s,v)$, $H_{s,t,v}=1$ denotes disagreement
between the simulated and observed real policy ranking, including a simulation
tie. We predict this label from the available scene-level quality score:
\begin{equation}
 \hat H_{s,t,v}=\mathbf{1}[Q_{s,v}<\tau_{-s,t}].
 \label{eq:quality-detector}
\end{equation}
A separate threshold for each task is selected on the other two scenes by
maximizing balanced accuracy, with sensitivity and then the smallest threshold
as tie-breakers. All variants of the target scene are held out together.
If training labels contain only one class, the predictor returns that class;
no quality threshold is learned in that fold. Target-scene real outcomes are
used only to score predictions. Test and OBS use identical eligible conditions
within each comparison. Test PSNR, SSIM, and LPIPS use the same threshold
selection and held-out conditions; LPIPS is negated so that higher scores
consistently indicate better quality. These three metrics use aligned
captured reference images for Test views. A second scene-level label requires all three task
rankings to agree with real evaluation. The available OBS scores are shared
scene--variant measurements, rather than separate task-specific scores.
\begin{figure*}[!t]
\centering
\includegraphics[width=\textwidth]{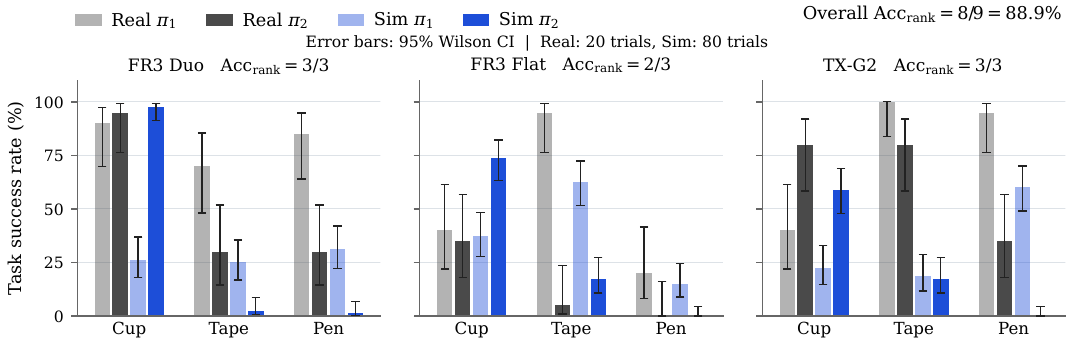}
\caption{\textbf{Policy ranking in real and simulated evaluation.}
Black/gray bars show real success and blue bars show simulated success;
light and dark bars denote $\pi_1$ and $\pi_2$, respectively, as in
Fig.~\ref{fig:teaser}. Policy labels are fixed within each task across scenes
and domains. Error bars are marginal 95\% Wilson confidence intervals from
20 real and 80 simulated trials per policy. The observed ranking agrees
in 8/9 scene--task pairs (88.9\%); Flat Cup is reversed.}
\label{fig:policy-success}
\end{figure*}
\section{Policy Ranking and Reconstruction Quality}
\label{sec:eval}

\textbf{Evaluation questions.}
The experiments answer four separate questions.  The clean experiment asks
whether simulation selects the same policy as the real robot.  The injected
artifact experiment asks whether reconstruction defects change that
selection.  The view-count experiment asks whether reducing capture support
produces similar visual defects and, when rollouts are run, changes policy
selection.  The quality experiment asks whether image-quality scores can
identify scenes whose policy rankings are unreliable.  Each question has a
separate outcome; a lower task success rate alone does not imply a ranking
failure.

\subsection{Clean policy ranking}
\label{sec:clean-results}
\textbf{Q: Can clean simulation reproduce which of $\pi_1$ and $\pi_2$ performs
better on the real robot?}

\textbf{A: Yes.  It reproduces the real policy ranking in 8 of 9
scene--task comparisons (88.9\%), with one inversion in Flat/Cup.}
The success-rate estimate, ranking metric, and secondary absolute-rate error
are defined in the protocol (Eqs.~\ref{eq:success-rate}--\ref{eq:rate-mae}).
Simulation ties and reversals are counted as failures of the choice and are
reported separately.

Fig.~\ref{fig:policy-success} compares real and simulated success rates for
the two policies on Cup, Tape, and Pen, and summarizes their rank agreement.
Eight of nine scene--task pairs
agree with real evaluation, giving 88.9\% pairwise ranking accuracy. The
single inversion is Flat Cup: real success is 40\% versus 35\%, while
simulated success is 37.5\% versus 73.75\% at 600 frames for the same policy labels.
Clean MAE is $31.3889$ percentage points
($0.3139$ on the unit-interval scale); absolute-rate error and
pairwise ranking agreement are distinct outcomes, and neither substitutes for the other.

\begin{table*}[t]
\centering
\caption{\textbf{Effect of each counterfactual (CF) condition on the nine scene--task policy comparisons.}
Significance: comparisons whose Fisher-test status ($p<0.05$) is lost or
gained relative to Clean; the others keep their status. Ranking vs.\ Clean: comparisons whose simulated
ranking is reversed or tied relative to Clean. Agreement with real: comparisons
whose simulated ranking matches the real ranking. Clean agrees in 8/9 (Flat Cup
is reversed) and no CF condition corrects it, so every reversal or tie is a lost agreement.}
\label{tab:cf-significance-ranking-transitions}
\footnotesize
\setlength{\tabcolsep}{5pt}
\renewcommand{\arraystretch}{0.92}
\begin{tabular}{lccccc}
\toprule
& \multicolumn{2}{c}{Significance} & \multicolumn{2}{c}{Ranking vs.\ Clean} & Agreement \\
\cmidrule(lr){2-3}\cmidrule(lr){4-5}\cmidrule(lr){6-6}
CF condition & Lost & Gained & Reversed & Tied & with real \\
\midrule
Hole 54$^\circ$ & 1 & 1 & 1 & 0 & 7/9 \\
Hole 90$^\circ$ & 4 & 0 & 1 & 0 & 7/9 \\
Floater low8 & 1 & 0 & 0 & 1 & 7/9 \\
Floater mid8 & 1 & 1 & 1 & 0 & 7/9 \\
Floater high8 & 1 & 0 & 1 & 0 & 7/9 \\
Warping & 2 & 1 & 0 & 0 & 8/9 \\
Central pinhole & 5 & 0 & 0 & 0 & 8/9 \\
Contiguous 30\% removal & 3 & 0 & 1 & 0 & 7/9 \\
Random 30\% removal & 3 & 0 & 2 & 0 & 6/9 \\
\bottomrule
\end{tabular}
\vspace{-6pt}
\end{table*}

\subsection{Counterfactual artifacts}
\label{sec:counterfactual}
\textbf{Q: How do injected reconstruction artifacts change policy-ranking
agreement and the significance of policy success-rate differences?}

\textbf{A: Holes and floaters reduce agreement from $8/9$ to $7/9$;
Warping preserves $8/9$ agreement, although significance status changes.}
Table~\ref{tab:cf-significance-ranking-transitions} evaluates all nine
scene--task comparisons for each condition, with 80 simulated trials per
policy. Transitions use the Clean reference results in
Sec.~\ref{sec:clean-results} as the common reference.
Both Hole angles and all three Floater configurations yield $7/9$
(77.8\%) ranking accuracy, compared with Clean's $8/9$ (88.9\%).
The additional disagreement is TX-G2 Tape: Hole $54^\circ$, Hole $90^\circ$,
and Floater mid8/high8 reverse its ranking, while Floater low8 makes the
policies tie. Warping preserves all Clean rankings and hence $8/9$ accuracy.

Ranking agreement and statistical significance capture different outcomes.
At the two-sided Fisher-test threshold $p<0.05$, Hole $90^\circ$ removes
significance in four comparisons: Duo Tape, Duo Pen, Flat Pen, and TX-G2 Pen.
For example, Duo Pen changes from $25/80$ versus $1/80$ in Clean
($p<0.001$) to $7/80$ versus $4/80$ ($p=0.534$), retaining its ranking.
Conversely, Hole $54^\circ$ reverses TX-G2 Tape from $15/80$ versus $14/80$
($p=1$) to $1/80$ versus $9/80$ ($p=0.018$), producing a significant
ranking opposite to the observed real ranking. Warping retains the TX-G2 Tape
ranking while making the difference significant ($14/80$ versus $5/80$,
$p=0.048$). Crossing the significance threshold is not itself a test that
the policy success-rate gap has changed significantly.

\subsection{Reconstruction from reduced image coverage}
\label{sec:disturbance}
\textbf{Q: How does reduced image coverage affect policy-ranking agreement?}

\textbf{A: Central pinhole retains $8/9$ agreement, while Contiguous 30\% removal
and Random 30\% removal reduce it to $7/9$ and $6/9$, respectively.}
As summarized in Table~\ref{tab:cf-significance-ranking-transitions},
Contiguous 30\% removal reverses TX-G2 Cup; Random 30\% removal reverses both TX-G2 Cup and TX-G2 Tape.
Random 30\% removal therefore has the largest observed accuracy decrease,
from 88.9\% to 66.7\% (22.2 percentage points). Central pinhole preserves
all Clean rankings but removes policy success-rate significance in five
comparisons, illustrating that unchanged ranking accuracy can coexist with
weaker statistical separation. Contiguous and Random 30\% removal each remove
significance in three comparisons. All reduced-coverage conditions use the same per-task time limits as Clean
(Sec.~\ref{sec:benchmark}).
\subsection{Quality response to view coverage}
\label{sec:nvs}
\textbf{Q: Does NVS-SQA reveal how each view-coverage disturbance changes
scene quality relative to clean reconstruction?}

\textbf{A: Weakly. The change is scene- and mode-dependent and does not
share a sign, but it is small against the spread between scenes.}\\
On held-out Test views, Contiguous 30\% removal moves mean NVS-SQA by $-0.060$ for
Duo, $+0.031$ for Flat, and $-0.018$ for TX-G2; Random 30\% removal by $+0.028$,
$+0.009$, and $-0.039$; Central pinhole by $+0.018$, $-0.005$, and
$+0.004$. Every one of these is far smaller than the $0.224$ spread between
the three scenes' own clean scores, so the metric separates scenes more
strongly than it separates a scene's capture conditions. The score-change
definition and the Test/OBS view protocols are specified in the benchmark
protocol.

\subsection{Scene verification across tasks and scenes}
\textbf{Q: Can scene-quality scores identify reliable policy rankings across
Cup, Tape, and Pen without target-scene real evaluation?}

\textbf{A: OBS improves Cup decisions, but this advantage does not extend
to Tape or to the aggregate current evaluation.}
We evaluate Cup, Tape, and Pen across three scenes and five matched variants:
Hole $54^\circ$/$90^\circ$, Central pinhole, Contiguous 30\% removal, and Random 30\% removal.
We compare NVS-SQA on Test/OBS with PSNR, SSIM, and LPIPS on Test,
selecting task-specific thresholds on two scenes and evaluating the third.
Table~\ref{tab:quality-screening} reports reliability-classification accuracy.
Among the five quality-based methods in Table~\ref{tab:quality-screening}, SSIM achieves the highest pooled task-wise accuracy (32/45), while Test NVS-SQA
performs best for deciding whether all three task rankings are reliable
(13/15). NVS-SQA on OBS improves Cup decisions but not aggregate accuracy
(29/45 versus 30/45 for Test). Pen's perfect score reflects constant
predictions because all its labels are reliable, rather than demonstrated
failure detection. The always-reliable predictor achieves $35/45$ on the
same conditions, exceeding all five quality-based methods in pooled accuracy.
All three Tape disagreements occur in TX-G2, so the TX-G2 fold trains on a single class and every quality-based method misses them; Tape accuracy differences therefore arise only from false alarms on agreement-labeled conditions.

\begin{table}[t]
\centering
\caption{Scene-held-out reliability classification: correct/total decisions
on 15 matched conditions per task. NVS: NVS-SQA; Total: pooled task
decisions; All-three: reliability of all three task rankings.}
\label{tab:quality-screening}
\footnotesize
\setlength{\tabcolsep}{2.5pt}
\begin{tabular}{lrrrrr}
\toprule
Metric/view & Cup & Tape & Pen & Total & All-three \\
\midrule
NVS Test & 7/15 & 8/15 & 15/15 & 30/45 & 13/15 \\
NVS OBS & 12/15 & 2/15 & 15/15 & 29/45 & 7/15 \\
PSNR Test & 8/15 & 8/15 & 15/15 & 31/45 & 9/15 \\
SSIM Test & 8/15 & 9/15 & 15/15 & 32/45 & 7/15 \\
LPIPS Test & 8/15 & 5/15 & 15/15 & 28/45 & 10/15 \\
\bottomrule
\end{tabular}
\end{table}

\section{Conclusion}
\label{sec:conclusion}
We presented YUBI-Real2Sim to evaluate whether reconstructed scenes support
reliable policy evaluation without real-robot verification in the target
scene. Clean simulation agrees with real policy rankings in $8/9$
scene--task comparisons. Agreement is $7/9$ under holes and floaters and
$6/9$ under Random 30\% removal; unchanged rankings can nevertheless lose
statistical separation between policies. Scene-held-out evaluation uses
quality scores to predict reliability from real labels in other scenes.
OBS-based NVS-SQA improves Cup decisions, but the benefit does not extend
to the aggregate three-task result. The results distinguish agreement with
real rankings, statistical separation of policies, and quality-based
prediction of reliability as separate evaluation outcomes.

\begin{acknowledgements}
This paper is based on results obtained from a project, JPNP25015, commissioned by the New Energy and Industrial Technology Development Organization (NEDO).
\end{acknowledgements}

\end{document}